\documentclass[twoside,twocolumn]{article}
\makeatletter
\def\section{\@startsection {section}{1}{\z@}{-2.0ex plus
    -0.5ex minus -.2ex}{1.5ex plus 0.3ex minus .2ex}{\large\bf\raggedright}}
\makeatother
\def\eqvsp{}

\usepackage{latexsym}
\usepackage{graphicx}
\usepackage{amsmath}
\def\,{\mskip 3mu} \def\>{\mskip 4mu plus 2mu minus 4mu} \def\;{\mskip 5mu plus 5mu} \def\!{\mskip-3mu}
\def\dispmuskip{\thinmuskip= 3mu plus 0mu minus 2mu \medmuskip=  4mu plus 2mu minus 2mu \thickmuskip=5mu plus 5mu minus 2mu}
\def\textmuskip{\thinmuskip= 0mu                    \medmuskip=  1mu plus 1mu minus 1mu \thickmuskip=2mu plus 3mu minus 1mu}
\textmuskip
\def\eqvsp{}
\def\beq{\eqvsp\dispmuskip\begin{equation}}    \def\eeq{\eqvsp\end{equation}\textmuskip}
\def\beqn{\eqvsp\dispmuskip\begin{displaymath}}\def\eeqn{\eqvsp\end{displaymath}\textmuskip}
\def\bqa{\eqvsp\dispmuskip\begin{eqnarray}}    \def\eqa{\eqvsp\end{eqnarray}\textmuskip}
\def\bqan{\eqvsp\dispmuskip\begin{eqnarray*}}  \def\eqan{\eqvsp\end{eqnarray*}\textmuskip}
\def\bmat#1{\left[ \begin{array}{#1}}
\def\emat{\end{array} \right]}

\def\myparskip{\vspace{1.3ex plus 0.5ex minus 0.5ex}\noindent}

\newtheorem{theorem}{Theorem}

\newtheorem{definition}[theorem]{Definition}
\newenvironment{proof}{\paradot{Proof}}{\qed\vskip 1ex}
\def\paradot#1{\myparskip{\bfseries\boldmath{#1.}}}

\def\nq{\hspace{-1em}}
\def\qed{\hspace*{\fill}$\Box\quad$\\}
\def\fr#1#2{{\textstyle{#1\over#2}}}
\def\req#1{(\ref{#1})}
\def\SetR{I\!\!R}

\def\qmbox#1{{\quad\mbox{#1}\quad}}

\def\v{\boldsymbol}

\def\a{\alpha}
\def\k{\kappa}
\def\minv{\overline}
\def\bino#1#2{{\textstyle\binom{#1}{#2}}}
\def\appr{\doteq}

\begin{document}

\title{\bf\huge\hrule height5pt \vskip 4mm
Equivalence of Probabilistic Tournament \\
and Polynomial Ranking Selection%
\vskip 4mm \hrule height2pt}

\author{
{\bf Kassel Hingee} \ -- \
\normalsize Department of Mathematics @
 Australian National University \ \ \ \ \ \ -- \
\texttt{hinkass@gmail.com}
\and
{\bf Marcus Hutter} --
\normalsize RSISE$\,$@$\,$ANU and SML$\,$@$\,$NICTA,
Canberra, ACT, 0200, Australia --
\texttt{marcus@hutter1.net}
}
\date{March 2008}

\maketitle

\begin{abstract}
\bf Crucial to an Evolutionary Algorithm's performance is its selection
scheme. We mathematically investigate the relation between
polynomial rank and probabilistic tournament methods which are
(respectively) generalisations of the popular linear ranking and
tournament selection schemes. We show that every probabilistic
tournament is equivalent to a unique polynomial rank scheme. In
fact, we derived explicit operators for translating between these
two types of selection. Of particular importance is that most linear
and most practical quadratic rank schemes are probabilistic
tournaments.
\end{abstract}

\section{Introduction}

\paradot{Evolutionary algorithms}
Evolutionary Algorithms (EAs) are probabilistic search algorithms
based on evolution \cite{Goldberg:89,Eiben:03}. They operate by
exploiting the information contained in a population  of possible
solutions (via similarities between individuals). The aim is to find
an individual that maximises (or minimises) an objective function,
which maps from individuals to the real line.
The population is transformed by first selecting individuals.
Mutation and/or recombination is then used to either replace a few
individuals from the population or create an entirely new
population.

\paradot{Standard selection methods}
The most prevalent methods for selecting individuals are
proportionate, linear rank, tournament, and truncation
\cite{Hutter:06fuo}. In proportionate selection individuals are
chosen with a probability proportional to their fitness (the value of
the objective function evaluated at the individual) \cite{Back1994}. A
common method to gain more control over selection pressure, is to scale the
fitness values before the selection is made \cite{Back1994}. Linear
ranking proceeds by ordering the population according to their
fitness. The chance that an individual is selected is then a linear
function of its (unique) rank \cite{Back1994}. Tournament selection
creates a tournament by randomly choosing $t$ individuals, the best
individual in the tournament is then selected. For truncation
selection the $k$ fittest individuals have uniform probability of
selection, while the remainder have zero chance of being selected.

The choice of selection scheme is crucial to algorithm performance.
If the selection pressure is too high then diversity of the
population decreases rapidly and the algorithm converges prematurely
to local optima or worse. With too little pressure there is not
enough push toward better individuals and the population takes too
long to converge. Many methods to choose or adapt the selection
pressure or avoid the problem otherwise have been invented (see
\cite{Hutter:06fuo} for some references). A particularly simple one
is fitness uniform selection, which uniformly selects a fitness
value, and then the individual with fitness closest to this value.

It is quite profitable to study selection schemes due to their
generality. They depend only on the set of fitness values and not on
the rest of the algorithm. Hence their behaviour can be studied in
isolation and the results applied to any evolutionary algorithm.
In this paper we introduce and study generalizations of rank and
tournament selection (both actually only depend on the rank and
not the absolute fitness value itself).

\paradot{Polynomial rank selection}
Linear ranking has a small range of selection pressures (from \cite{Back1994}, for a population of $n$ individuals the probability that the fittest individual is selected must be between $1/n$ and $2/n$),
but it has the flexibility of a real-valued parameter that can vary continuously (the slope
of the linear function). Ranking schemes
with high selection pressures, such as when the probability of
selection is an exponential function of the rank, have occasionally been
used \cite{Wieczorek2002}. It is natural then to generalise from linear to
polynomial functions to cover the instances where medium pressure is
required. Hence the probability of an individual with rank $k$ being
selected with a polynomial rank scheme of degree $d$ is:
\beq\label{eqPoly1}
  P(I=k)=\sum_{l=1}^{d+1} a_l k^{l-1}
\eeq
where $a_l\in \SetR$ are parameters defined by the algorithm
designer.
For simplicity we assume that selection is performed with
replacement and each individual has unique rank, however our results
still hold when there are ties in the rank. The only restriction on
the $a_k$ is that they must produce a proper probability
distribution, i.e.\ for a population of $n$ individuals: $P(I=k)\geq
0$ for all $k=1,2,...,n$ and $\sum_{k=1}^n P(I=k)=1$. Hence, while
the population is ordered, the schemes may favour low ranks, high
ranks or neither, depending on the choice of the $(a_l)$.

This selection method encompasses the low pressures of linear
schemes ($a_3=...=a_{d+1}=0$) and can give good approximations of
the high pressure exponential cases (via Taylor polynomials).
Furthermore the wealth of general knowledge about polynomials means
that while it has numerous parameters (coefficients of the
monomials), it is also easy to predict their impact.

\paradot{Probabilistic tournament selection}
Tournament selection has a large range \cite{Back1994}, but a
discrete parameter, leaving the possible selection pressures
somewhat restricted. This can be overcome by selecting
probabilistically from the tournament, rather than always choosing
the best in the tournament. However the extra parameters required
are not easy to understand. Their precise effect on the behaviour is
not at all obvious. Probabilistic Tournament selection still only
sorts $t\ll n$ individuals, making it much faster than any ranking
scheme.

Let $i_s$ be the (rank of the) individual in position
$s\in\{1,...,t\}$ of the rank-ordered tournament. We call $s$ the
\emph{seed} of $i$. Let $P(I_s=k)$ be the probability that seed $s$
has rank $k$. In any given tournament, the probability that the seed
$s$ individual is chosen will be a user defined constant $\a_s$.
Then the probability of an individual $k$ being selected through a
size $t$ probabilistic tournament is:
\beq\label{eqTourn1}
  P(I=k)=\sum_{s=1}^t \a_s P(I_s=k)
\eeq
Standard (deterministic) tournament always selects the individual of
highest rank in the tournament, i.e.\ $\a_1=1$ and $\a_2=..=\a_t=0$.

To ensure that choosing a winner from the tournament makes sense,
the $\a_s$ must satisfy the probability constraints $\a_s\geq
0\,\forall s$ and $\sum_{s=1}^t \a_s =1$. We assume that the
tournament is created by random selection \emph{with} replacement
and for now that each individual in the population has a unique
fitness. This defines $P(I_s=k)$ (Section \ref{secPST}). Note that
even if every individual in the population is unique, it is possible
for it to be repeated in the tournament.

\paradot{Previous work on the relation between rank and tournament selection}
In this paper we investigate the equivalence between the generalised
schemes \req{eqPoly1} and \req{eqTourn1} with the aim of providing a
scheme that combines the superior understanding of polynomial rank
with the speed of probabilistic tournament.

B\"{a}ck \cite{Back1994} found that an individual's chance of
selection in deterministic tournament selection is a polynomial,
hence each is equivalent to a polynomial rank selection method.
Wieczorek and Czech \cite{Wieczorek2002}, and Blickle
\cite{Blickle1995CA} arrived at the same conclusion using a
different method. So while the name `polynomial rank selection' is
new, its concept is fairly old.

The study of probabilistic tournaments isn't new either: Hutter
\cite[p.11]{Hutter:91cfs} proved that every size $2$ probabilistic
tournament is a linear rank scheme, and Goldberg \cite{Goldberg:91}
did the same but only for a continuous population.
Fogel \cite{Fogel:1988} applied to the traveling salesman problem, a
variation wherein each individual underwent numerous t=2
tournaments. The probability of winning each tournament was
dependent on the fitness of the individuals involved and the
individuals selected were those with the highest number of wins.

\paradot{Contents: Equivalence of polynomial rank and probabilistic tournament selection}
We extend these results by finding that every $t$ sized
probabilistic tournament is equivalent to a polynomial rank scheme
with a polynomial degree of $d=t-1$ or less (Section \ref{secPST}).
We continue on to show that the equivalence is unique (Section
\ref{secUnique}), and give an explicit expression for the inverse
map (Section \ref{secPtoT}). This allows the establishment of simple
criteria for polynomial rank schemes that are probabilistic
tournaments (Section \ref{secPisT}). Unfortunately not every
possible polynomial rank scheme satisfies the criteria, but most
(and in the limit of an infinite population, all) linear and most
``interesting'' quadratic ones are equivalent to probabilistic
tournaments. This is good enough for all practical purposes, if it
generalises to higher order polynomials.

\paradot{Notation}
Throughout the paper we use the following notation. If not otherwise
indicated, an index has the full range as defined in this table.

\begin{tabbing}
\hspace{2cm}         \= \hspace{11cm} \= \kill
{\bf Symbol }        \> {\bf Explanation}   \\[0.5ex]
$\delta_{ij}$        \> Kronecker symbol \\
                     \> ($\delta_{ij}=1$ for $i=j$ and $\delta_{ij}=0$ for $i\neq j$) \\[0.5ex]
$n$                  \> Number of individuals in the population\\
$i,j,k$              \> Rank (unique label) of individuals $\in\{1,...,n\}$ \\[0.5ex]
$\iota,\k$       \> Rank indices that only run from $1,...,t$\\
$p,q,r,s$            \> Seed index $\in\{1,...,t\}$\\
$I_s$                \> Rank of the individual with seed $s$\\
$I$                  \> Rank of the individual selected\\
$\pi_i=P(I=i)$       \> Probability that $i$ is selected \\[0.5ex]
$l$                  \> Polynomial coefficients index $\in\{1,...,d+1\}$\\[0.5ex]
$a_l$                \> Coefficients of $x^{l-1}$ for the polynomial\\
$\a_1, \a_2,...,\a_t$ \> Tournament selection coefficients\\
$x_i, \v x, (x_i)$   \> Vector $\v x=(x_i)=(x_1,...,x_t)$ \\[0.5ex]
$\Delta_m$           \> $m-1$ dimensional probability simplex \\[0.5ex]
\end{tabbing}

\section{Probability of Selection via a Tournament}\label{secPST}

In this section we find the probability of an individual being
successful (the winner) via tournament selection. This will provide
a formula for an equivalent ranking selection scheme. It is
sufficient to consider just one selection event in isolation, since
we consider selection with replacement.

We assume a population ${\cal P}$ consisting of $n$ individuals
$c_1,...,c_n$ with fitness $f_1,...,f_n$. Without loss of generality
we assume that they are ordered, i.e.\ $f(i)\geq f(j)$ for all $j\leq
i$. For now we also assume that all fitness values are different,
hence individual $c_i$ has rank $i$. The rank is all we need in the
following, and we will say ``individual $i$'', meaning ``individual
$c_i$''.

\begin{definition}[polynomial rank selection]\it
Polynomial $\v a$-ranking selects individual $c_k$ from population
$\cal P$ with probability $P(I=k)=\sum_{l=1}^{d+1}a_l k^{l-1}$
\end{definition}

\begin{definition}[probabilistic tournament selection]\it
A probabilistic $\v\a$-tournament selects $t$ individuals from
population $\cal P$ uniformly at random with replacement. Let
$c_{I_s}$ be the individual of rank $s$ in the tournament, called
seed $s$ (while it has rank $I_s$ in the population). Finally the
seed $s$ individual, $I_s$, is chosen with probability $\a_s=P(I=I_s)$
as the winner $I$.
\end{definition}

\begin{theorem}[tournament=polynomial]\label{thmPST}\it
Probabilistic $\v\a$-tournament selection coincides with polynomial
$\v a$-ranking (for $d=t-1$ and suitable $\v a$).
\end{theorem}

\begin{proof}
We derive an explicit expression for the probability $\pi_k$ that
the tournament winner has rank $k$. Any seed $s$ may have rank $k$
($I_s=k$) and may be the winner ($I=I_s$), hence
\beqn
  \pi_k \;\equiv\; P(I=k) \\
  \;=\; \sum_{s=1}^t P(I=I_s)P(I_s=k) \;=\; \req{eqTourn1}
\eeqn
where we have exploited that by definition the probability that $I=I_s$ is
independent of the rank $I_s=k$. $P(I_s=k)$ is the probability that
seed $s$ has rank $k$. It is difficult to formally derive an
expression for $P(I_s=k)$, but we can easily get it by considering
distribution functions. The probability of an individual selected
into the tournament having a particular rank is $1/n$, hence having
rank equal to or less than $k$ is $k/n$ and larger than $k$ is $1-k/n$.
Further, $I_r\leq k \wedge I_{r+1}>k$ if and only if $r$ seeds have
rank $\leq k$ and $t-r$ seeds have rank $>k$, hence
\beqn
  P(I_r \leq k \wedge I_{r+1}>k) \;=\;
  \bino{t}{r} (\fr{k}{n})^r(1\!-\!\fr{k}{n})^{t-r}
\eeqn
since there are $\bino{t}{r}$ ways of choosing $r$ individuals with
rank $\leq k$ from $t$ individuals. The above expression is a
polynomial in $k$ of degree $t$. Together with
\beqn
  P(I_s\leq k) \;=\; \sum_{r=s}^t P(I_r \leq k \wedge I_{r+1}>k),
\eeqn
we get the explicit expression
\bqa\label{eqPIs}
  & & P(I_s=k) \;=\; P(I_s\leq k)-P(I_s\leq k-1)
\\ \nonumber
  & & =\; \sum_{r=s}^t \bino{t}{r}
           \big[(\fr{k}{n})^r(1\!-\!\fr{k}{n})^{t-r}
              - (\fr{k-1}{n})^r(1\!-\!\fr{k-1}{n})^{t-r}\big]
\eqa
Using the binomial theorem to find the $k^t$ and $k^{t-1}$
coefficients in the square brackets above reveals that the former
coefficients cancel out while the latter do not. This implies that $P(I_s=k)$
is a polynomial in $k$ of degree (at most) $t-1$, and thus the weighted average \req{eqTourn1} is as well.
Summing \req{eqTourn1} over the population yields $\sum_{k=1}^n
P(I=k)=1$, as it should, since the tournament coefficients are such
that some individual is always chosen.
Consequently, every tournament is a polynomial rank scheme of degree
at most $t-1$ (one can choose $\v\a$ such that it is of lower
degree).
\end{proof}

\begin{figure}
\setlength{\unitlength}{1mm}
\begin{picture}(89,78)(-4,-1)
\put(0,0){\includegraphics[width=85mm]{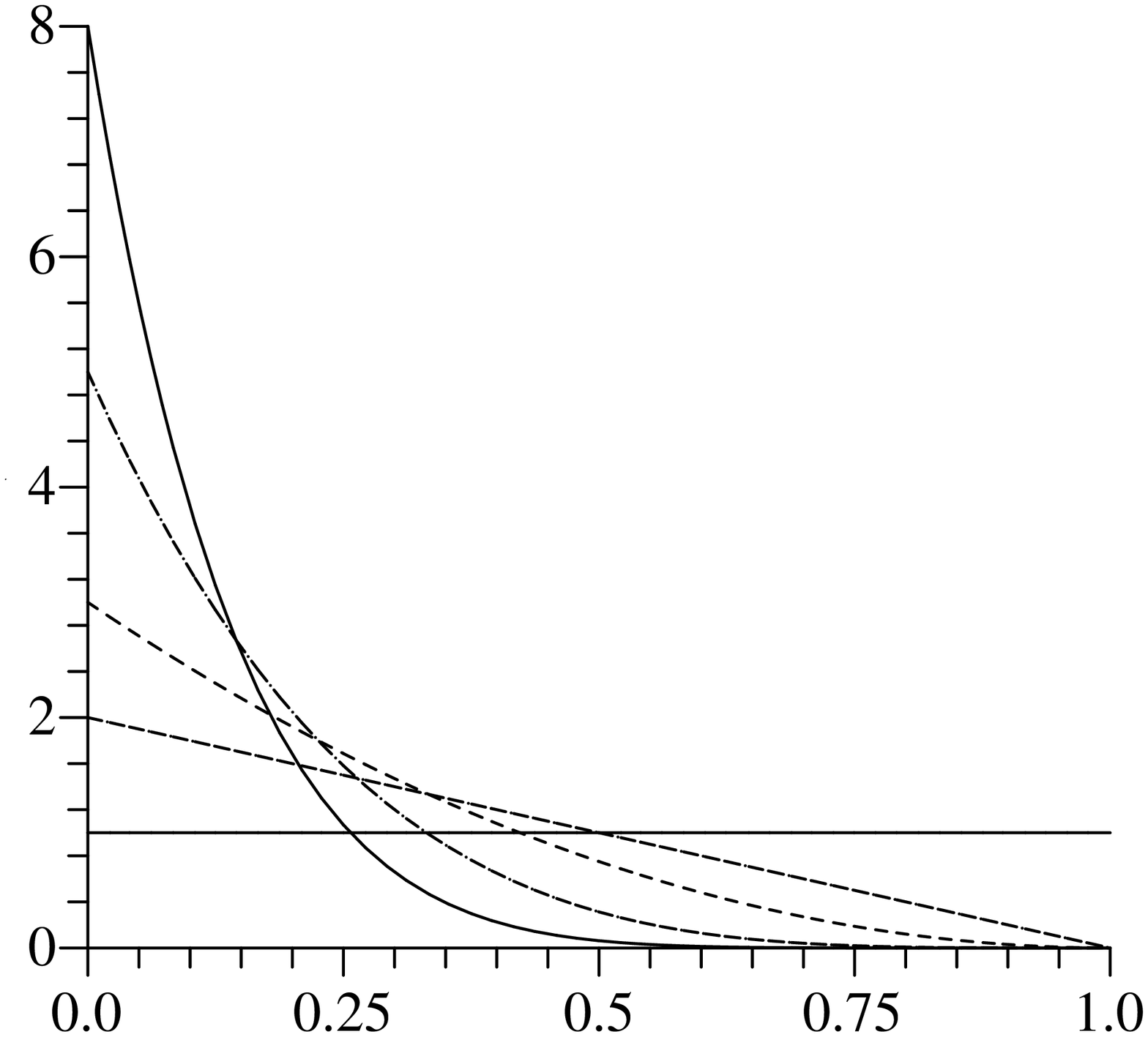}}
\put(-4,15.5){\large$t=1$}
\put(6.5,20){\large$t=2$}
\put(-4,31.5){\large$t=3$}
\put(-4,48){\large$t=5$}
\put(8,65){\large$t=8$}
\put(40,-1){\Large$x$\normalsize}
\end{picture}
\caption{\label{figtourndet}[tournament probabilities for large $n$] \it
Probability density $nP(I_1=xn)$ that the tournament winner has rank $xn$,
for tournament size $t=1,2,3,5,8$.}
\end{figure}

\paradot{Examples}
Expression \req{eqPIs} can be rewritten as
\beqn
  P(I_s=k) \;=\; \sum_{r=0}^{s-1} \bino{t}{r}
    \big[(\fr{k-1}{n})^r(1\!-\!\fr{k-1}{n})^{t-r}
       - (\fr{k}{n})^r(1\!-\!\fr{k}{n})^{t-r} \big]
\eeqn
which will be convenient in the following examples.
Standard tournament always selects $I_1$ ($\a_1=1$), hence \cite{Back1994}
\beqn
  P(I=k) \;=\; P(I_1=k) \;=\; (1-\fr{k-1}{n})^t - (1-\fr{k}{n})^t
\eeqn
See Figure \ref{figtourndet}.
For $t=1$ there is no selection pressure, $P(I=k)=\fr1n$.
For $t=2$ we get
\beqn
  P(I_1=k) = \fr{2n-2k+1}{n^2} \qmbox{and}
  P(I_2=k) = \fr{2k-1}{n^2}
\eeqn
Hence probabilistic tournaments of size 2 lead to
linear ranking \cite{Hutter:91cfs}
\bqa\nonumber
  & & \nq\nq P(I\!=\!k) \;=\; \a_1 P(I_1\!=\!k)+\a_2 P(I_2\!=\!k) \;=\; a_1+a_2 k,
\\ \label{PIkt2}
  & & \nq\nq a_1=\fr1{n^2}[(2n+1)\a_1-\a_2], \qquad a_2 \;=\; \fr2{n^2}(\a_2-\a_1)
\eqa

\paradot{Remark}
More interesting is actually the converse, replacing rank selections
by equivalent efficient tournaments. Before we can answer this, we
need to break down \req{eqPIs} into a product of simple regular
matrices.

\section{The Map from Tournament to Polynomial is Unique}\label{secUnique}

The next natural question is whether different tournament bias
$\v\a$ implies different selection probability. It seems plausible
that the maps from tournaments $\v\a$ to rank probabilities $\v\pi$
and to polynomial coefficients $\v a$ are injective, but the proof
is fairly involved. The good news is that construction in the proof
allows us to find a closed form expression for the desired inverse.
Let $\Delta_m=\{\v x\in\SetR^m:x_i\geq 0\,\forall i,\sum_{i=1}^m
x_i=1\}$ be the $m-1$ dimensional probability simplex, i.e.\
$\v\pi\in\Delta_n$ and $\v\a\in\Delta_t$.

\begin{theorem}[tournament$\rightarrow$polynomial]\label{thmUnique}\it
The function $R:\Delta_t\to\Delta_n$ in \req{eqTourn1}, mapping
tournament probabilities $\v\a$ to rank probabilities $\v\pi$, is
total, linear, and injective:
\beqn
  \pi_k=P(I=k)=\sum_{s=1}^t R_k^s\a_s, \qmbox{i.e.} \v\pi=R\v\a,
\eeqn
where $R_k^s=P(I_s=k)$ is defined in \req{eqPIs}. Matrix $R$ can
also be written as a product $R=\minv D P F C D = V N F C D = V T$
with matrices $\minv D$, $P$, $F$, $C$, $D$, $V$, and $N$ defined in
\req{defVinv}, \req{defP}, \req{defF}, \req{defC}, \req{defV},
\req{defQ}, and \req{defN}.
Similarly, the function $T:\Delta_t\to\SetR^t$, mapping $\v\a$ to
polynomial coefficients $\v a$, is unique, linear and injective:
\beqn
  a_l=\sum_{s=1}^t T_l^s\a_s, \qmbox{i.e.} \v a=T\v\a,
\eeqn
where matrix $T = N F C D$.
\end{theorem}

\begin{proof}
Tournament always selects one individual from $\cal P$ as the
winner, hence $R\v\a\in\Delta_n$ for every $\v\a\in\Delta_t$. See
the proof of Theorem \ref{thmPST} for how to prove this formally.

\paradot{Matrices $H$ and $G$}
We now prove injectivity. With
\beqn
  H_k^r := (\fr{k}{n})^r(1-\fr{k}{n})^{t-r}  \qmbox{and}
  G_k^r := \bino{t}{r}(H_k^r-H_{k-1}^r)
\eeqn
we can write \req{eqPIs} as
\beq\label{eqRsumG}
  R_k^s \;\equiv\; P(I_s=k) \;=\; \sum_{r=s}^t G_k^r
\eeq

\paradot{Einstein notation}
Einstein's sum convention will be convenient in the following
argument: When an index occurs repeatedly in the multiplication of
two objects, a sum over the index over its full range is implicitly
understood, e.g.\ $G_k^r D_r^s$ means $\sum_{r=1}^t G_k^r D_r^s$.

\paradot{Lower-triangular matrix $D$}
The lower-triangular matrix
\beq\label{defV}
  D_r^s:=\left\{ \begin{array}{rcl} 1 & \mbox{if} & s\leq r\\
                0 & \mbox{if} & s>r \end{array} \right.
\eeq
has the property that $\sum_{r=1}^t G_k^r D_r^s = \sum_{r=s}^t
G_k^r$. Using Einstein's sum convention this allows us to rewrite
\req{eqRsumG} as
\beqn
  R_k^s = G_k^r D_r^s
\eeqn
i.e.\ as a product of an $n\times t$ matrix $G$ with a $t\times t$
matrix $D$.

\paradot{Inverse of $D$}
The ``inverse'' of $D$ is:
\beq\label{defVinv}
  \minv D_k^i
  \;:=\; \left\{ \begin{array}{rl}
   1 & \mbox{if } k=i\\
  -1 & \mbox{if } i=k-1\\
   0 &\mbox{otherwise}\end{array}
  \right\} \;=\; \delta_{k,i}-\delta_{k-1,i}
\eeq
This is a matrix with $1$ on the primary diagonal; $-1$ on the
diagonal that is below the primary diagonal; and $0$ otherwise.

\paradot{Decomposing $G$}
$G_k^r$ itself can actually be decomposed into $\minv D_k^i$ and
$H_i^q$ and a pure diagonal matrix
\beq\label{defC}
  C_q^r = \bino{t}{q}\delta_{q,r}
\eeq
comprised of the binomial coefficients:
\beqn
  G_k^r \;=\; (H_k^q-H_{k-1}^q)C_q^r
        \;=\; \minv D_k^i H_i^q C_q^r
\eeqn
(note that $\minv D$ is the inverse of an $n\times n$ sized D matrix
here).

\paradot{Decomposing $H$ into $P$ and $F$}
We can decompose $H_i^q$ further be using the binomial identity:
\bqan
  H_i^q &=& (\fr{i}{n})^q(1-\fr{i}{n})^{t-q} \\
  &=& \textstyle (\fr{i}{n})^q\sum_{s=1}^{t-q}\bino{t-q}{s}(\fr{-i}{n})^{t-q-s} \\
  &=& \textstyle \sum_{s=1}^{t-q}(-)^{t-q-s}\bino{t-q}{s}(\fr{i}{n})^{t-s}\\
  &=& \textstyle \sum_{p=q}^t(-)^{p-q}\bino{t-q}{t-p}(\fr{i}{n})^p
\eqan
So $H_i^q=P_i^p F_p^q$, where $P$ is a matrix
of monomials:
\beq\label{defP}
  P_i^p:=(\fr{i}{n})^p,
\eeq
and $F$ is a lower-triangular matrix composed of various binomials:
\beq\label{defF}
  F_p^q:=\left\{ \begin{array}{cl}
   (-)^{p-q}\bino{t-q}{t-p} & \mbox{if } q\leq p \\
                0 & \mbox{otherwise} \end{array}\right.
\eeq
\paradot{Matrices $N$, $V$, and $R$} 
Putting everything together we have
\beqn
  R_k^s \;=\; \minv D_k^i P_i^p F_p^q C_q^r D_r^s
\eeqn
%
The (linear) map $R_k^s$ is a polynomial in $k$ of degree (at most)
$t-1$. We can find its coefficients by rewriting
\bqa\label{VPeqQN}
  \minv D_k^i P_i^p
  &=& P_k^p-P_{k-1}^p
  \;=\; (\fr{k}{n})^p-(\fr{k-1}{n})^p
\\ \nonumber
  &=& \sum_{l=1}^p k^{l-1}(-)^{p-l}\bino{p}{l-1}(\fr1n)^p
  \;=\; V_k^l N_l^p
\eqa
where
\bqa\label{defQ}
  V_k^l &:=& k^{l-1}, \qmbox{and}
\\ \label{defN}
N_l^p &:=& \left\{\begin{array} {cr} (-)^{p-l}\bino{p}{l-1}(\fr1n)^p & l\leq p\\
0 & \mbox{otherwise}\end{array} \right.
\eqa
Hence we get the alternative representation
\beq\label{eqRalt}
  \pi_k=R_k^s\a_s \;=\; V_k^l N_l^p F_p^q C_q^r D_r^s\a_s
\eeq

\paradot{Injective}
Matrices $D$, $\minv D$, and $F$ are lower-triangular matrices with
1 in the diagonal, and hence are invertible (thus injective). $C$ is diagonal and $N$
upper triangular, both nowhere zero on the diagonal, hence
invertible too. The first $t$ rows of $V$ map from a set of $t$
coefficients $\v b$ to the polynomial $p(x)=\sum_{l=1}^t b_l
x^{l-1}$ evaluated at $x=1,2,3,...,t$. A degree $t-1$ polynomial
like $p$ is uniquely determined by $t$ image points
(see Appendix),
hence $V$ is injective. Similarly for P or exploit $P_k^l=k
V_k^l(\fr1n)^l$ (no summation). This proves that $R$ is injective.

\paradot{Matrix $T$}
Combining the map from $\v a$ to $\v\pi$
\beqn
  \pi_k \equiv P(I=k)
  = \sum_{l=1}^t a_l k^{l-1}
  = V_k^l a_l
  \qmbox{i.e.} \v\pi=V\v a,
\eeqn
with $a_l=T_l^s\a_s$ we get
\beqn
  \pi_k = V_k^l T_l^s\a_s
\eeqn
Comparing this with \req{eqRalt} and using injectivity of $V$ we see
that
\beq\label{defT}
  T_l^s \;=\; N_l^p F_p^q C_q^r D_r^s
\eeq
which is injective, since $N$, $F$, $C$, and $D$ are invertible.
\end{proof}

\paradot{Discussion}
Given a Polynomial Rank scheme it is possible and easy (using
computer software) to find if it is equivalent to a probabilistic
tournament (and get the corresponding parameters) by applying the
inverse of $T$ to $\v a$. If the output satisfies the probability
requirements $\v\a\in\Delta_t$, then it is indeed a probabilistic
tournament.

\section{Map from Polynomial Ranking to Tournaments}\label{secPtoT}

We now derive explicit expressions for the really interesting
converse of map $T$, which allows replacement of inefficient
rank selections by equivalent efficient tournaments.
From the last section we know that the inverse exists.

\begin{theorem}[polynomial$\rightarrow$tournament]\label{thmTtoP}\it
The function $\minv T:\SetR^t\to\SetR^t$,
mapping polynomial coefficients $\v a$ to
tournament parameters $\v\a$ is linear
\beqn
  \a_s=\sum_{l=1}^t\minv T_s^l a_l, \qmbox{i.e.} \v\a=\minv T\v a
\eeqn
where matrix $\minv T=T^{-1}=\minv D\,\minv C\,\minv F\,\minv N$, with
$\minv D$, $\minv C$, $\minv F$, $\minv N$
defined in \req{defVinv}, \req{defCinv},
\req{defFinv}, \req{defNinv}.
$\v a$-polynomial ranking
can be implemented as an $\v\a$-tournament if and only if, $\v\a=\minv T\v a\in\Delta_t$.
\end{theorem}

\paradot{Inverse matrices}
In the following $P$ and $V$ respectively denote the upper $t\times t$ submatrix
of $P$ and $V$. The inverse matrices are as follows
\bqa\label{defCinv}
  \minv C_r^q &:=& \delta_{r,q}/\bino{t}{r}
\\ \label{defFinv}
  \minv F_q^r &:=& \bino{t-r}{t-q} \mbox{ if } r\leq q \mbox{ and 0 else}
\\ \label{defNinv}
  \minv N_p^l &=& \minv P_p^\iota D_\iota^\k V_\k^l
\\ \label{defPinv}
  \minv P_l^\k &:=& n^l\minv V_l^\k\fr1\k \qquad\mbox{(no summation)}
\eqa
The inverse of the diagonal matrix $C$ is obvious.
The expression for $\minv P$ immediately follows from
$P_\k^l=\k V_\k^l(\fr1n)^l$ (no summation).

$F_p^q\minv F_q^r=0$ for $r>p$ (since then either $r>q$ or $q>p$) and
for $r\leq p$ we have
\bqan
  F_p^q\minv F_q^r
  &=& \textstyle\sum_{q=r}^p (-)^{p-q}\bino{t-q}{t-p}\bino{t-r}{t-q}
\\
  &=& \textstyle\bino{t-r}{t-p}\sum_{q=r}^p \bino{p-r}{q-r}(-)^{p-q}
  \;=\; \delta_{p,r}
\eqan
The first equality is by definition, the second equality is a simple
reshuffling of factorials, and the last equality follows from the
well-known binomial identity $\sum_{i=0}^m(-)^i\binom{m}{i}=0$ for
$m\geq 1$. This proves that $\minv F$ is the inverse of $F$.

Unfortunately we were not able to invert $N$ directly, although $N$
seems similar to (the transpose of) $F$. So we used relation
\req{VPeqQN} to invert $N$ in \req{defNinv}. But now we need the
inverse of $P$, which can be reduced by \req{defPinv} to the inverse
of $V$.

\paradot{Inverse of $V$}
The most difficult matrix to invert is $V$. This special Vandermonde
matrix $V$ can be written as a product of a lower $L$ and
upper-triangular matrix $U$, whose inverses are \cite{Turner:66}:
\bqan\label{defQinv}
  \minv V_l^\k &:=& \minv U_l^s \minv L_s^\k
\\
  \minv L_s^\k &:=& {(-)^{s-\k}\over (s-\k)!(\k-1)!} \mbox{ for } s\geq \k \mbox{ and 0 else}
\\
  \minv U_l^s &:=& S_s^{(l)} \mbox{ = Stirling numbers of the first kind}
\eqan
The Stirling numbers $S_s^{(l)}$ numbers are defined as the coefficients of the
polynomial $x(x-1)...(x-s+1)$, i.e.\ by
\beqn
  \sum_{l=0}^s S_s^{(l)} x^l = {x!\over(x-s)!} \qmbox{and} S_s^{(l)}=0 \mbox{ for } l>s
\eeqn
There are many ways to compute $S_s^{(l)}$, e.g.\ recursively by
$S_{s+1}^{(l)}=S_s^{(l-1)}-s S_s^{(l)}$ or directly
\cite[p.824]{Abramowitz:74}.
For $r\geq \k$ we get
\bqan
  V_r^l\minv V_l^\k
  &=& \sum_{l=1}^t r^{l-1}\sum_{s=\k}^l
    {S_s^{(l)}(-)^{s-\k}\over (s-\k)!(\k-1)!}
\\
  &=& \sum_{s=\k}^t {[\sum_{l=1}^s r^{l-1}S_s^{(l)}] (-)^{s-\k}\over (s-\k)!(\k-1)!}
\\
  &=& \sum_{s=\k}^t {[(r-1)...(r-s+1)] (-)^{s-\k}\over (s-\k)!(\k-1)!}
\\
  &=& \bino{r-1}{\k-1}\sum_{s=\k}^r\bino{r-\k}{r-s}(-)^{s-\k} \;=\; \delta_{\k,r}
\eqan
The case $r<\k$ is similar. This shows that $\minv V$ is the inverse of
(the first $t$ rows of) $V$.

\paradot{Linear ranking example}
For $t=d+1=2$ we can compute the matrices by hand. This list of
(reduced) matrices is a useful sanity check for the reader's
own implementation:
\beqn
\begin{array}{ll}
        F = \bino{\;\;1\,0}{-1\,1},\;\;
  \minv F = \bino{1\,0}{1\;1}, & \nq
        H = \fr1{n^2}\bino{n-1\,2n-4\,3n-9\;\cdots\;\; 0}{\quad 1\quad\;\; 4\quad\;\; 9\quad\cdots\;\; n^2}\!^\top
\\[3pt]
        C = \bino{2\;0}{0\,1},\quad
  \minv C = \fr12\bino{1\,0}{0\,2}, &
        G = \fr1{n^2}\bino{n-1\,n-3\,\cdots\,-1+n}{\quad 1\quad 3\quad\cdots\;2n-1}\!^\top
\\[3pt]
        N = \fr1{n^2}\bino{n\,-1}{0\;\;2}, &
  \minv N = \fr{n}2\bino{2\;1}{0\,n}
\\[3pt]
        P = \fr1{n^2}\bino{n\,2n\,3n\,\cdots\,n^2}{1\;\;4\;\;9\;\,\cdots\,n^2}\!^\top, &
  \minv P = \fr{n}2\bino{\;\;4\;\;-1}{-2n\;\;n}
\\[3pt]
        V = \bino{1\,1\,\cdots\,1}{1\,2\,\cdots\,n}\!^\top, &
  \minv V = \bino{\;\;2\,-1}{-1\;\;1}
\\[3pt]
        T = \fr1{n^2}\bino{2n+1\,-1}{-2\quad 2}, &
  \minv T = \fr{n}4\bino{2\quad 1\quad\!}{2\,2n+1}
\\[3pt]
  \minv U = \bino{1\,-1}{0\;\;1},\quad
  \minv L = \bino{\;\;1\,0}{-1\,1}, &
        R = \fr1{n^2}\bino{2n-1\,2n-3\,\cdots\quad 1\quad}{\quad 1\quad\; 3\;\;\cdots\;2n-1}\!^\top
\end{array}
\eeqn
We see that $\v\pi=R\v\a$ and $\v a=T\v\a$ coincide with
\req{PIkt2}, as they should.

\paradot{Computational complexity}
Together this allows us to compute $\v\a$ from $\v a$ and vice versa
in time $O(t^2)$ and $\v\pi$ from $\v\a$ in time $O(nt)$. Once
$\v\a$ is known, tournament selection needs only time $O(t)$ per winner
selection.

\section{What Polynomial Selection Schemes are Tournaments?}\label{secPisT}
Theorem \ref{thmTtoP} does not give us conditions under which the
resulting tournament parameters $\v\a=\minv T\v a$ are valid. We look for such conditions so that we can reliably change/create tournament schemes in the more understandable set of polynomial rank schemes. Without these conditions there can be no guarantee that whatever created would be a probabilistic tournament.

\paradot{Range of linear ranking}
Let us first consider the case of linear ranking ($d=1$),
\beqn
  P(I=k)=a_1+a_2 k
\eeqn
We want to find the range of $a_1$ and $a_2$
for which this is a proper probability distribution in $\Delta_n$.
The sum-constraint leads to
\beqn
  1 = \sum_{k=1}^n P(I=k)
    = a_1 n+a_2 \fr12 n(n+1)
\eeqn
\beq\label{a1froma2}
  \Longrightarrow\quad
  a_1=\fr1n[1-\fr12 a_2(n^2+n)]
\eeq
Next are the positivity constraints $P(I=k)\geq 0\,\forall k$.
A linear function is $\geq 0$ if and only if it is $\geq 0$ at its
ends, i.e.\ $P(I=1) \geq 0$ and $P(I=n)\geq0$.
Inserting \req{a1froma2} into these constraints yields:
\bqan
  P(I=1) &\equiv& a_1+a_2 \;\;\geq 0 \quad\iff\quad a_2 \;\leq\; \fr{2}{n^2-n}
\\
  P(I=n) &\equiv& a_1+a_2n \geq 0 \quad\iff\quad a_2 \geq -\fr{2}{n^2-n}
\eqan
So the possible linear rank schemes are those with
\beq
  |a_2|\leq \fr{2}{n^2-n}
  \qmbox{and} a_1 \mbox{ satisfying \req{a1froma2}} \label{LinRang}
\eeq

\paradot{Range of tournament size 2}
The example \req{PIkt2} shows that size $t=2$ probabilistic
$\v\a$-tournaments have $a_2=2(\a_2-\a_1)/n^2$. Since $\v\a\in\Delta_2$,
$a_2$ has range $-\fr2{n^2}...\fr2{n^2}$. As it should be, this is a
subset of the possible linear rank schemes. Hence the linear rankings
that are probabilistic tournaments are those with
\beq
  |a_2| \;\leq\; \fr2{n^2}
  \qmbox{and} a_1 \mbox{ given by \req{a1froma2}} \label{BinRang}
\eeq
This is slightly narrower than $|a_2|\leq\fr2{n^2-n}$, i.e.\ there
are some rankings that are not probabilistic tournaments. On the
other hand, $\fr2{n^2-n}/\fr2{n^2}$ tends to 1 as $n$ grows, hence
for $n$ large (e.g.\ about 100) nearly all linear rankings can be
translated into probabilistic tournaments. The coverage
is good enough for all practical purposes.

\paradot{The general case}
A probabilistic selection scheme is completely determined by
$\v\pi$, different $\v\pi$ correspond to different selection
schemes, and every $\v\pi\in\Delta_n$ is a valid selection scheme.
Hence, $\Delta_n$ is the set of all possible probabilistic selection
schemes.
The set of (valid) size $t$ tournament schemes is
\beqn
  R\Delta_t \;:=\; \{\v\pi=R\v\a:\v\a\in\Delta_t\} \;\subset\; \Delta_n
\eeqn
Since $R$ is injective, this is a $t-1$ dimensional irregular
simplex embedded in the $n-1$ dimensional simplex $\Delta_n$.

The set of (incl.\ invalid) degree (up to) $t-1$ polynomial ranking schemes is
\beqn
  V\SetR^t \;:=\; \{\v\pi=V\v a:\v a\in\SetR^t\} \;\not\subseteq\; \Delta_n
\eeqn
This is a $t$-dimensional hyperplane. Only $\v\pi$ in $\Delta_n$ are
valid, hence $V\SetR^t\cap\Delta_n$ is the set of (valid) polynomial
ranking schemes. The intersection of a simplex with a plane gives a
closed, bounded, convex polytope, in our case of dimension $t-1$.
The Krein-Milman Theorem \cite[p.707]{Edwards1965} says that for a
closed, bounded, convex subset $A$ of $\SetR^t$ with a finite number
of extreme points (=corners), $A$ is the convex hull of the extreme
points of $A$. Hence the extreme points of $V\SetR^t\cap\Delta_n$
completely characterize/define the set.

If/since we are not concerned with the covering of
$V\SetR^t\cap\Delta_n$ in $\Delta_n$ itself, we can study the covering
in the lower-dimensional polynomial coefficient space $\SetR^t$.
The set=polytope of all polynomial coefficients $\v a$ that lead to valid
selection probabilities is
\beqn
  \minv V\Delta_n \;:=\; \{\v a\in\SetR^t: V\v a\in\Delta_n\}
\eeqn
while the set=simplex of coefficients reachable by tournaments is
\beqn
  T\Delta_t \;:=\; \{ \v a=T\v\a: \v\a\in\Delta_t\} \;\subset\; \minv V\Delta_n
\eeqn

These sets are the images of $V\SetR^t\cap\Delta_n$ and the simplex
$\Delta_t$ under $\minv V$ and $T$ respectively. These maps are
injective (Section \ref{secPtoT}) so  $\minv V\Delta_n$ and
$T\Delta_t$ are completely determined by their extreme points. The
extreme points of $\Delta_t$ are just the conventional $\SetR^t$
basis vectors $\v e_s$, so $T\Delta_t$ is the convex hull of $\{T(\v
e_s):s=1,...,t\}$. The polytope $V\SetR^t\cap\Delta_n$ can be quite
complex, and finding the extreme points daunting. This is
essentially what we did for the $t=2$ case in the above paragraphs.

We estimated the proportion of degree $t-1$ polynomials covered by
$T\Delta_t$ for various $t$ using a Monte-Carlo
algorithm\footnote{The $t=2$ case was calculated directly from
\req{LinRang} and \req{BinRang}}   (Table \ref{AreaCov}). It shows
that for $n\geq100$, practically all linear rank schemes are
probabilistic tournaments.

Nothing concrete can be concluded about the coverage for $t=4,5$.
Table \ref{AreaCov} only suggests that the number of $t-1$ degree
polynomials equivalent to $t$-sized tournaments decreases as $t$
increases.

\paradot{Tournament size 3}
In the $t=3$ case we can extend our knowledge by finding $\minv
V\Delta_n$ graphically. The restriction  $\sum_{k=1}^n \pi_k=1$
means that the $k^2$ coefficient $a_3$, is completely determined by
$a_1$ and $a_2$.
\beq a_3=\frac{1}{\sum_{k=1}^n k^2}\left(1-n a_1 -\fr12 n(n+1)a_2\right)
  \label{t3plane}
\eeq
Hence $\minv V \Delta_n$ is a 2 dimensional hyperplane. $P(I=k)\geq
0$ for each $k$ defines a set of halfspaces:
$\{(a_1,a_2,a_3):a_1+a_2 k+a_3 k^2\geq0\}$; $\minv V\Delta_n$ is
their intersection (over $k$)  restricted to the plane given by
\req{t3plane}.

\begin{figure}
\setlength{\unitlength}{1.1mm}
\begin{picture}(80,74)(-4,0)
\put(0,0){\includegraphics[width=90.2mm]{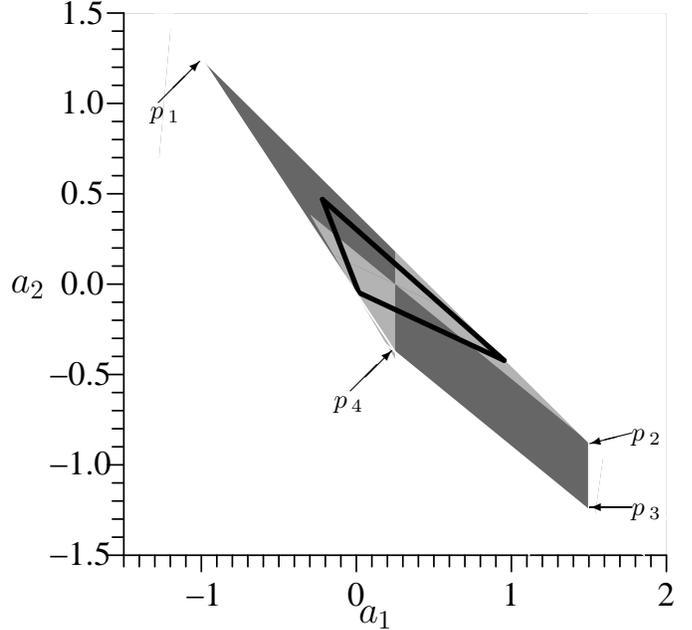}}
\put(12.8,61){$p\,_1$}
\put(13.8,63){\vector(1,1){5}}
\put(71,22){$p\,_2$}
\put(71,23){\vector(-4,-1){5}}
\put(71,13){$p\,_3$}
\put(71,14){\vector(-4,0){5}}
\put(35,26){$p\,_4$}
\put(37,28){\vector(1,1){5}}
\put(38,0){\Large$a_1$\normalsize}
\put(-4,40){\Large$a_2$\normalsize}
\end{picture}
\caption{\label{pic3}[$n=4,t=3$] \it The shaded region is the set of possible
polynomials, whilst the light grey area is the set of the most useful polynomials. The triangle is the boundary of the set
that can be written as $t=3$ tournaments. At $p_1$: $a_3\appr -0.246$. At $p_2$: $a_3\appr 0.159$. At $p_3$: $a_3\appr 0.236$. At $p_4$: $a_3\appr 0.023$. }
\end{figure}

\begin{figure}
\setlength{\unitlength}{1mm}
\begin{picture}(87,79)(-5,-3)
\put(0,0){\includegraphics[width=82mm]{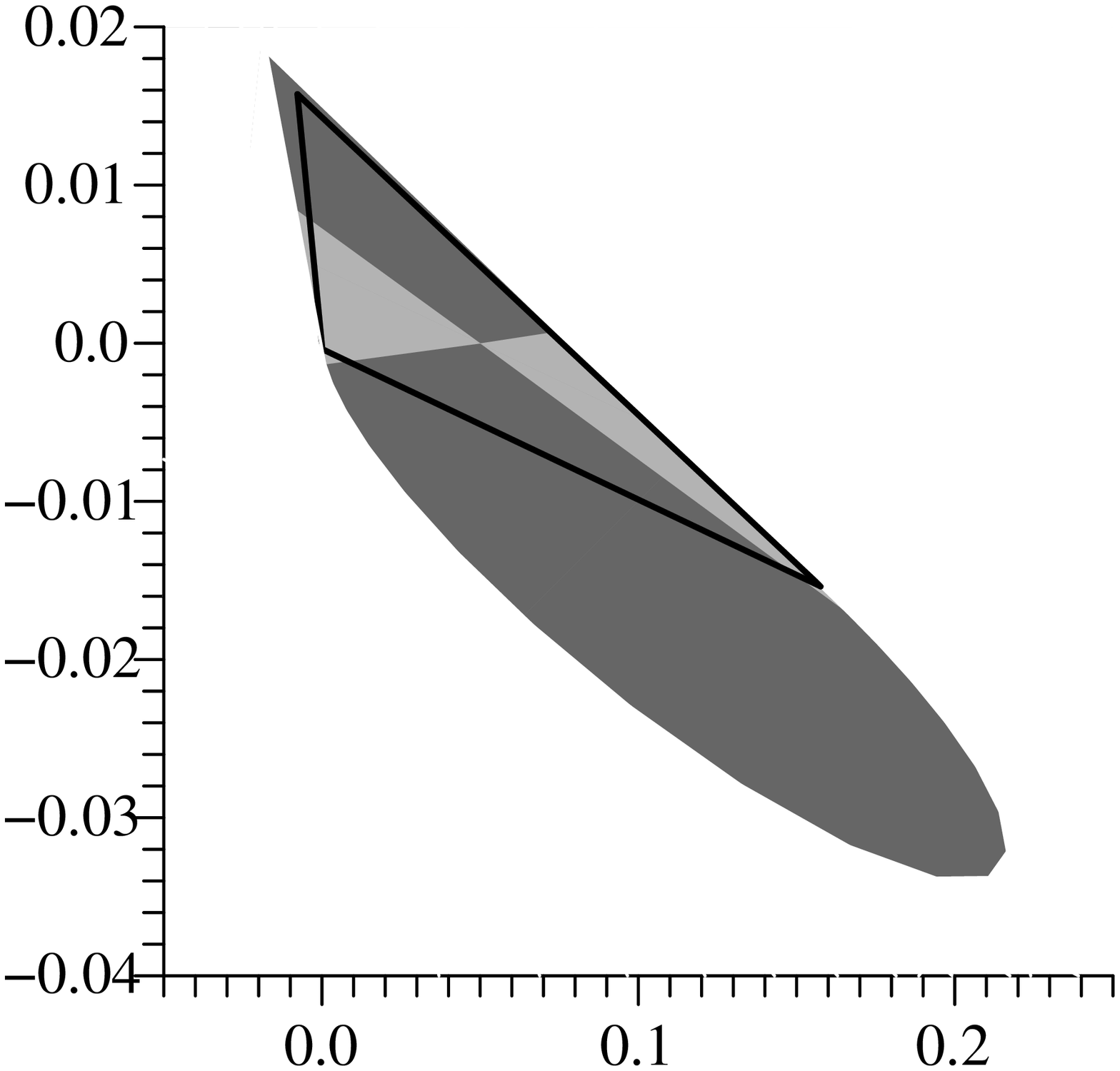}}
\put(12,64){$p\,_1$}
\put(13,66){\vector(1,1){5}}
\put(73.5,12.5){$p\,_2$}
\put(73.5,13.5){\vector(-4,1){5}}
\put(14.5,48){$p\,_3$}
\put(16.5,50){\vector(4,1){5}}
\put(41,-3){\Large$a_1$\normalsize}
\put(-5,40){\Large$a_2$\normalsize}
\end{picture}
\caption{\label{pic20}[$n=20$,$t=3$] \it The shaded region is the set of possible
polynomials, whilst the light grey area is the set of the most useful polynomials. The triangle is the boundary of the set
that can be written as $t=3$ tournaments. At $p_1$: $a_3\appr -8.56 \times 10^{-4}$. At $p_2$: $a_3\appr 1.08\times 10^{-3}$. At $p_3$: $a_3\appr 1.91\times 10^{-4}$.}
\end{figure}

\begin{figure}
\setlength{\unitlength}{1.05mm}
\begin{picture}(85,81)(-3,-2)
\put(0,0){\includegraphics[width=86.1mm]{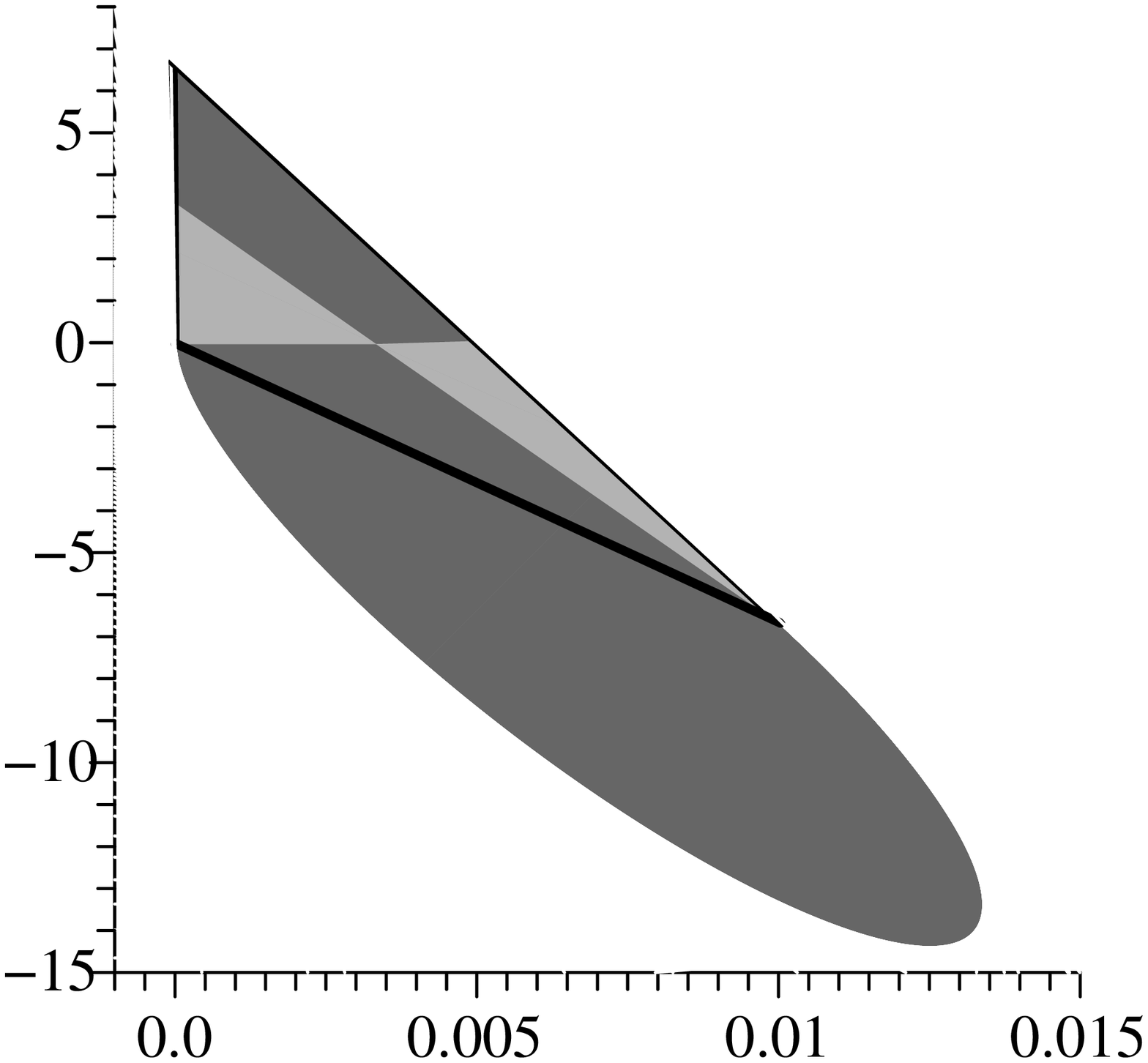}}
\put(6,75){\large $\times 10^{-5}$\normalsize}
\put(18,71){$p\,_1$}
\put(17.5,71){\vector(-4,-1){5}}
\put(73,10){$p\,_2$}
\put(73,11){\vector(-4,0){6}}
\put(9,44.2){$p\,_3$}
\put(10,46.2){\vector(1,2){2}}
\put(38,-2){\Large$a_1$\normalsize}
\put(-3,40){\Large$a_2$\normalsize}
\end{picture}

\caption{\label{pic300}[$n=300$,$t=3$] \it The shaded region is the set of possible
polynomials, whilst the light grey area is the set of the most useful polynomials. The triangle is the boundary of the set
that can be written as $t=3$ tournaments. At $p_1$: $a_3\appr -2.18\times 10^{-7}$. At $p_2$: $a_3\appr 3.53\times 10^{-7}$. At $p_3$: $a_3\appr 1.21\times 10^{-7}$.}
\end{figure}

$T\Delta_3$ is simply a filled triangle with corners $\{T(\v
e_s):s=1,2,3\}$.  Comparison with $\minv V\Delta_n$ (Figures
\ref{pic3}, \ref{pic20} and \ref{pic300})
suggests that the coverage of $T\Delta_3$ is stable for
$n\rightarrow \infty$. Hence for large populations about a third of
the quadratic polynomials can be written as size-$3$ probabilistic
tournaments.

In practice, selection schemes with probability monotonically
increasing with fitness are used. So not the whole of $\minv
V\Delta_n$ is interesting, but only the subset of monotonically
increasing or possibly decreasing probabilities on $\{1,2,...,n\}$
(light grey in figures \ref{pic3}, \ref{pic20} and \ref{pic300}).
The remainder of  $\minv V\Delta_n$ is composed of schemes
that favour the middle ranks or both high and low
ranked individuals (dark grey).

Any polynomial scheme $P(I=k)=a_1+a_2 k+\frac{1}{\sum_{i=1}^n
i^2}\left(1-n a_1 -\fr12 n(n+1)a_2\right)k^2$ is a
parabola\footnote{We temporarily consider k to range over the real
line}, so it is symmetric about it's stationary point, $x_{st.pt.}$.
Hence $P(I=k)$ is monotonic on $\{1,2,...,n\}$ if and only if
$x_{st.pt.}$ lies outside the interval $(1+\frac12,n-\frac12)$.

i.e.
\beqn x_{st.pt.}=\frac{-a_2}{1-n a_1 -\fr12
  n(n+1)a_2}\left(\sum_{i=1}^n i^2\right) \leq 1+\frac12
\eeqn
\emph{OR}
\beqn
  x_{st.pt.}\geq n-\frac12
\eeqn

Figure \ref{pic300} suggests that these regions of usefulness
effectively lie entirely in $T\Delta_3$ for $n\geq 300$. Hence for
sufficiently large $n$ the most useful degree 2 polynomial schemes
are perfectly reproduced by some probabilistic tournament.

\begin{figure}
\includegraphics[width=\columnwidth]{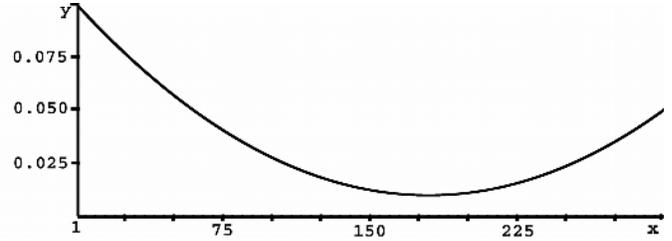}
\caption{\label{notinT} [n=300] The polynomial $y=0.01-10^{-4}x+2.781\times 10^{-7}x^2$. This is an example of a usable quadratic polynomial that is not equivalent to a probabilistic tournament.}
\end{figure}

An example of a less applicable selection scheme is the polynomial
given by $a_1=0.01$ and $a_2=-1\times 10^{-4}$ (which lies in the
dark grey region). It favours both high ranks and low ranks (Figure
\ref{notinT}) and any algorithm using this scheme will spend half of
the time searching in the wrong place. However it is still usable
(like in fitness uniform selection \cite{Hutter:06fuo}).

The points $p_1$, $p_2$ ... are extreme points of $\minv V\Delta_n$.
They indicate that the range of $a_3$ values is significantly
smaller than the range of $a_2$ (which in turn has a smaller range
than $a_1$).

$\minv V\Delta_n$ being the intersection of a finite number of
halfspaces and planes means its boundary is actually a series of
straight lines. $\minv V\Delta_n$ appears curved in figures
\ref{pic20} and \ref{pic300} simply due to the many halfspaces that
are involved.

\begin{table}
\beqn
\begin{array}{|c|c|c|c|c|c|}
\hline
 & \mathbf{n=4} & \mathbf{n=10} & \mathbf{n=20} & \mathbf{n=100} & \mathbf{n=300}\\ \hline
 \mathbf{t=2} & 0.7500 & 0.9000 & 0.9500 & 0.9900 & 0.9967\\\hline
\mathbf{t=3} & 0.270& & 0.348 &0.342&0.332\\ \hline
\mathbf{t=4}&  & 0.12 & 0.15 & 0.16 &\\ \hline
\mathbf{t=5} & &0.02 & & &\\ \hline
\end{array}
\eeqn
\caption{fraction of possible $t-1$ degree polynomials that can be represented as $t$-sized probabilistic tournaments}
\label{AreaCov}
\end{table}

\section{Discussion/Conclusions}

\paradot{Rank ties}
Individuals with the same fitness lead to ties in the ranking.
If we break ties (arbitrarily but consistently), our theorems still
apply. The disadvantage is that the selection probability for two
individuals with the same fitness may not be the same. We can fix
this problem by breaking ties (uniformly) at random.
For instance, given a population of 3 individuals with two of them
having the same fitness, this results in effective selection
probabilities $\pi_1^{e\!f\!f}=\pi_1$ and
$\pi_2^{e\!f\!f}=\pi_3^{e\!f\!f}=\fr12(\pi_2+\pi_3)$.

\paradot{Further work}
Investigation of the set of possible polynomials with degree $d\geq
3$ will be helpful for those applications requiring higher selective
pressures. Furthermore, finding the proportion that are equivalent
to probabilistic tournaments may provide a reliable method for
making high-degree polynomial rank schemes more efficient.

Tournaments of size $t\ll n$ are significantly faster than ranking
schemes, so it would be beneficial to obtain a thorough
understanding of how many polynomial rank schemes are equivalent to
$t>d+1$ sized probabilistic tournaments.

\paradot{Conclusion}
We have found a strong connection between polynomial ranking and
probabilistic tournament selection.

We derived an explicit operator (\ref{defT}) that maps any
probabilistic tournament to its equivalent polynomial ranking
scheme, which is unique and always exists. Polynomial rank schemes
thus encompass linear ranking and deterministic (normal) tournament
selection, leaving designers with one less selection method (but
more parameters) to worry about.

Unfortunately, turning polynomial rank schemes into equivalent
probabilistic tournaments is not so straightforward. Only about a
third of the possible quadratic polynomials can be written as
size-$3$ probabilistic tournaments.

However, nearly all linear rank schemes have an equivalent size-$2$
probabilistic tournament. Hence nearly all can be made faster by
simply rewriting the scheme as a probabilistic tournament.

Furthermore, almost all the practical quadratic polynomials are
equivalent to some $t=2$ tournament. This is a good indication for
the investigation of $t>3$.

\appendix
\section{Appendix}

\paradot{Uniqueness of a polynomial given $t$ image points}
Let $\v\pi\in\SetR^t$ be the vector of $t$ image points
$\pi_\k=p(x_\k)$ for some $x_1,...,x_t$ of a polynomial
$p(x)=\sum_{l=1}^t a_l x^{l-1}$ with coefficient vector $\v
a\in\SetR^t$. In particular we have
\beqn
  \pi_\k \;=\; p(x_\k) \;=\; \sum_{l=1}^t a_l V_\k^l,
  \qmbox{where} V_\k^l \;=\; x_\k^{l-1}
\eeqn
If matrix $V$ is invertible, the polynomial (coefficients) would
be uniquely defined by $\v a=V^{-1}\v\pi$, which is what we set out to prove.
We now show that $V$ is invertible. Define the $t$
polynomials of degree $t-1$
\beqn
   p_s(x) \;=\;  \prod_{r=1\atop r\neq s}^t{x-x_r\over x_s-x_r}
   \;=\; \sum_{l=1}^t A_l^s x^{l-1}
\eeqn
Expanding the product in the numerator defines the coefficients $A_l^s$.
On $x_\k$ we get
\beqn
  \delta_{s \k} = p_s(x_\k) = \sum_{l=1}^n V_\k^l A_l^s
\eeqn
hence $A$ is the inverse of $V$. By explicitly expanding
$\prod(x-x_r)$ one can get an explicit expression for $A_l^s$, which
is unfortunately pretty useless.


\end{document}